\documentclass{article}

     \usepackage[nonatbib]{neurips_2020}

\usepackage[utf8]{inputenc} 
\usepackage[T1]{fontenc}    
\usepackage{hyperref}       
\usepackage{url}            
\usepackage{booktabs}       
\usepackage{amsfonts}       
\usepackage{nicefrac}       
\usepackage{microtype}      
\usepackage{graphicx}

\title{Experimenting with Self-Supervision using Rotation Prediction for Image Captioning}
\author{%
  David S.~Hippocampus\thanks{Use footnote for providing further information
    about author (webpage, alternative address)---\emph{not} for acknowledging
    funding agencies.} \\
  Department of Computer Science\\
  Cranberry-Lemon University\\
  Pittsburgh, PA 15213 \\
  \texttt{hippo@cs.cranberry-lemon.edu} \\
}

\begin{document}
\maketitle
\begin{abstract}


Image captioning is a task in the field of Artificial Intelligence that merges between computer vision and natural language processing. It is responsible for generating legends that describe images, and has various applications like descriptions used by assistive technology or indexing images (for search engines for instance). This makes it a crucial topic in AI that is undergoing a lot of research.
This task however, like many others, is trained on large images labeled via human annotation, which can be very cumbersome: it needs manual effort, both financial and temporal costs, it is error-prone and potentially difficult to execute in some cases (e.g. medical images).


To mitigate the need for labels, we attempt to use self-supervised learning, a type of learning where models use the data contained within the images themselves as labels.
It is challenging to accomplish though, since the task is two-fold: the images and captions come from two different modalities and usually handled by different types of networks. It is thus not obvious what a completely self-supervised solution would look like. How it would achieve captioning in a comparable way to how self-supervision is applied today on image recognition tasks is still an ongoing research topic. 


In this project, we are using an encoder-decoder architecture where the encoder is a convolutional neural network (CNN) trained on OpenImages dataset and learns image features in a self-supervised fashion using the rotation pretext task. The decoder is a Long Short-Term Memory (LSTM), and it is trained, along within the image captioning model, on MS COCO dataset and is responsible of generating captions.
Our GitHub repository can be found: \url{https://github.com/elhagry1/SSL_ImageCaptioning_RotationPrediction}

\end{abstract}

\section{Related Work}


Self-supervised learning can be broken down into two tasks: 

\textbf{A pretext task} where no labels are involved and where the model learns image features and finds information in the images that it can use as labels to carry out its mission (e.g. classification). 

\textbf{A downstream task} where small-labeled data is used (i.e. only a small percentage of the dataset has labels) to fine-tune the model and perform its initially intended purpose; image captioning in our case.

Generating image captions in a self-supervised way doesn't seem to be a prevalent topic in the existing literature, and we're suspecting that it comes from its difficulty rather than a lack of interest. The work that we have come across seems to be using either an unsupervised or a semi-supervised learning approach. While unsupervised learning deals with no supervision, semi-supervised learning merges between labeled and unlabeled data. We will be presenting some of the work in these two categories for a lack of self-supervised image captioning papers. 

\cite{gu} use semi-supervised learning by assuming that they have an image-caption corpus in a different language that they can use to generate captions with before translating them to a target language. In their paper, the source target is Chinese and English is the target one. Chinese here is called a pivot-language, and the task of performing a task through a proxy language, language pivoting. This strategy is used already in machine translation when a language is rich in resources and is used to bridge the cap between two languages that don't have a lot of data connecting them. 

\cite{gu} come up with a framework that is composed of an image captioner and a translation model, trained on distinct datasets. The image captioner, called \textit{image-to-pivot} generates the captions and passes them to the translation model, called \textit{pivot-to-target}, which translates the captions to the intended language.

Although language pivoting in the context of image captioning is a creative idea that can definitely be of very good use to languages that don't have image-caption datasets yet, it might not be very useful to English for instance since it relies on the assumption of an image captioning dataset being available in some other language. It is known however, that the biggest image-captioning datasets in English still fall a bit short due to captioning being a tedious annotation task and them introducing only a small subset of objects that are encountered in real-life. 

Additionally, machine-translation is known to have its own share of errors, and using it in conjunction with captioning could multiply these errors and accumulate them. This solution is therefore great for reusing data across languages but it does not seem to improve much on the current technology in resource-rich languages.

\cite{chen} bring the attention to how real world applications might not benefit much from the existing image-captioning models due to them being trained on "general" datasets. Indeed, in domain-specific settings, a well-performing model might not achieve satisfying results. For that reason, \cite{chen} use adversarial training to remedy to the issue using two different discriminators. One of them focuses on the domain and decided whether the captions generated are discernible from target domain sentences. The second one is responsible for assessing whether an image and caption are a good match. The two take the usual alternate training with the caption generator, each one of them trying to either distinguish more or generate better captions.

We believe that this approach is very useful and any advances towards more domain-specific and real-world scenarios hint to promising results in the future.
        
        
    The aim in \cite{feng} is not to use any image-captioning dataset, and to instead have separate sentence and image corpora. Adversarial learning is first used to generate sentences that can't be told apart from the sentences in the corpus. After that, information from a visual concept detector is used to reward based on words in the caption that are related to that visual concept, thus attaching the image concept to its caption.
    
    \cite{feng} use a common latent space between images and captions which motivates the semantic consistency of the sentences. An image feature can help in decoding a caption, and this latter can be used for the reconstruction of the former.

\section{Methods}

\subsection{Self-supervised Learning Methods}
        \cite{jigsaw} 
        The majority of conventional approaches can be classified as either generative or discriminative. Generic methods learn to produce or model pixels in the input space in a variety of ways. Pixel-level generation, on the other hand, is computationally expensive and may not be needed for representation learning. Discriminative methods train networks to execute pretext tasks using objective functions close to those found in supervised learning, but the inputs and marks are obtained from an unlabeled dataset.
        
        Unsupervised learning approaches can be divided into three categories: probabilistic, direct mapping (autoencoders), and manifold learning. Variables in a network are divided into measured and latent variables using probabilistic techniques. After that, learning is linked to evaluating model parameters that optimize the probability of latent variables given the observations. Unfortunately, as several layers are present, these models become intractable and are not optimized to deliver features efficiently. 
        
        The direct mapping method, which uses autoencoders, focuses on the latter part. Autoencoders define the feature extraction function (encoder) and the mapping from the feature to the input in a parametric manner (decoder). The direct mappings are learned by minimizing the autoencoder's reconstruction error between the input and the output.

        Manifold learning methods may be used if the data structure indicates the data points are likely to cluster along a manifold. This representation enables smooth factor variations to be directly mapped to smooth observation variations. Manifold learning strategies have a range of drawbacks, including the need to compute nearest neighbors, which scales quadratically with the number of samples. Besides also the need for a sufficiently large density of samples across the manifold, that becomes more challenging to obtain for high-dimensional manifolds.

        \subsubsection{Rotation}
        
        \cite{rot} uses CNNs to learn image features by recognizing the rotations applied to the images. The intuition behind it is that similar to a human, the CNN will need to understand the image concept and what objects appear in order to recognize the rotation. In particular, the encoder needs to localize objects that are salient, identify which orientation they have and what relations they have with their surroundings.
        
        \begin{figure}
            \includegraphics[width=\textwidth]{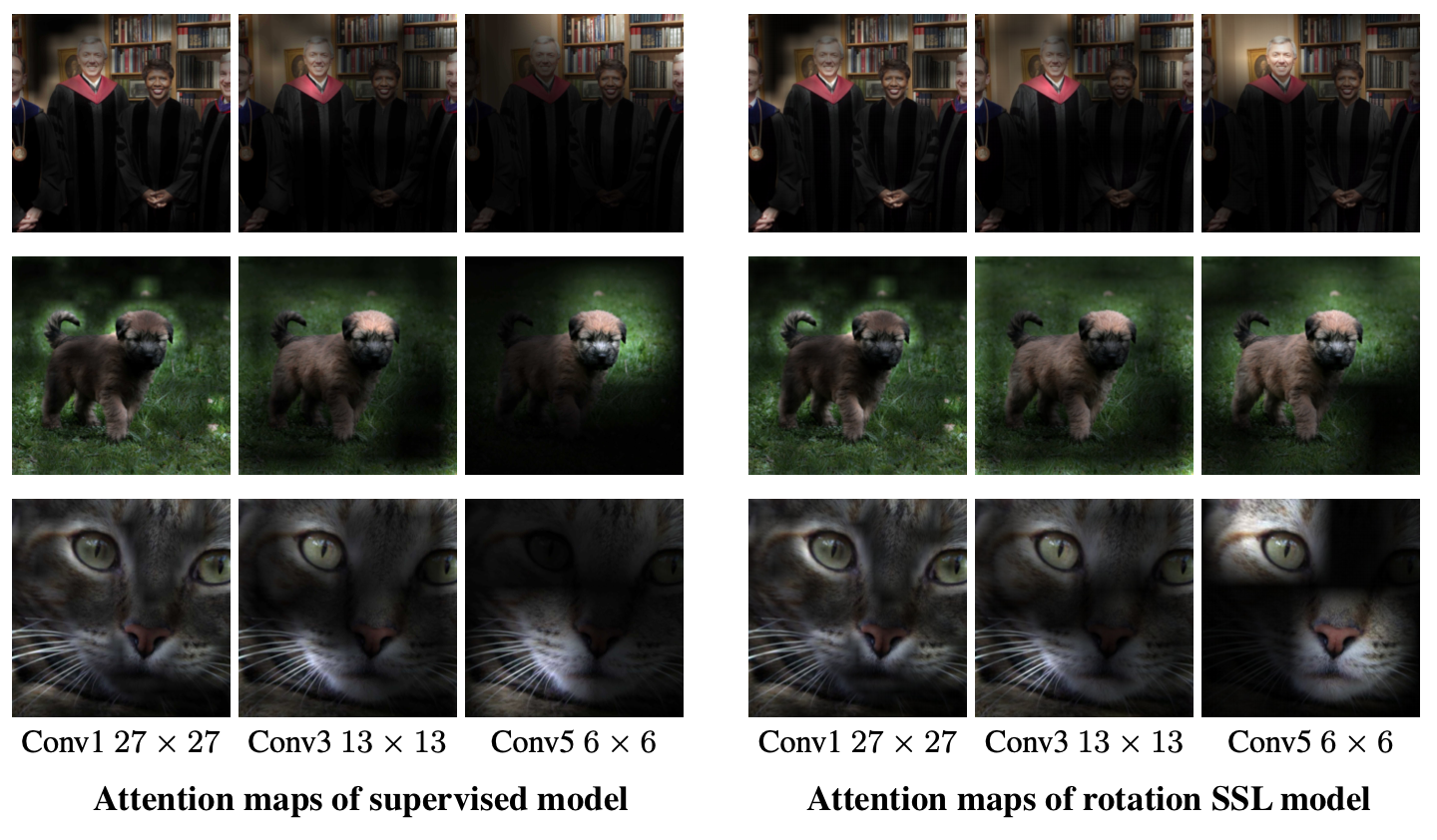}
            \caption{Attention maps from AlexNet supervised versus self-supervised using rotations.}
            \label{maps}
        \end{figure}
        
\begin{figure}
            \centering
            \includegraphics[width=0.4\textwidth]{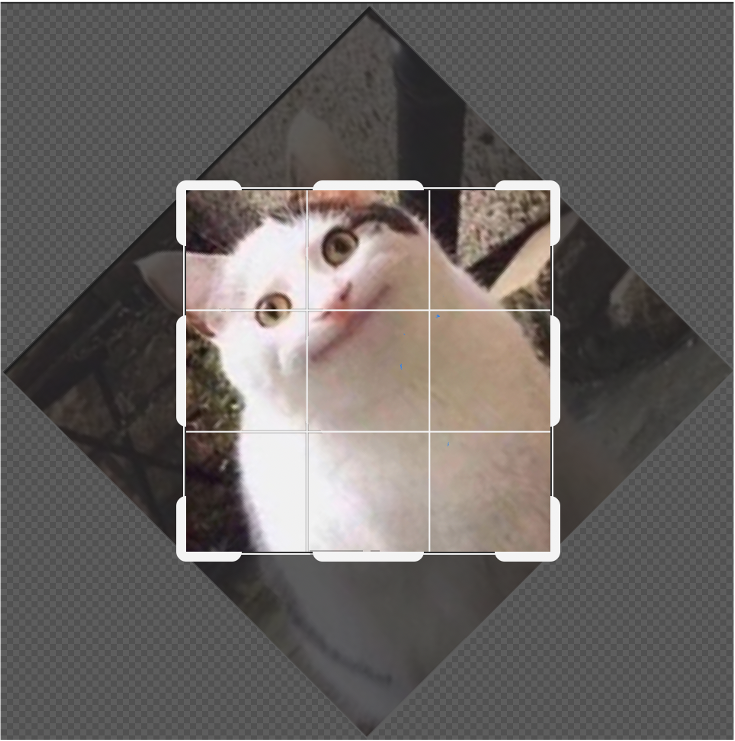}
            \caption{Data is lost when performing a rotation other than the four main ones (e.g. 45º)}
            \label{cat}
        \end{figure}

        This results in similar attention maps to those of a supervised learning model (Fig. \ref{maps}). This proves that although the idea is simple, it does achieve great results
        4 rotations are used: 0º, 90º, 180º and 270º. They're preferred as they're very easily implemented using transpose and flip operations which prevents commonly faced transformation issues where low-level artifacts are left and can distract the CNN from the important features. We also find them to be optimal since any other rotation will result in some of the data being lost as depicted in (Fig. \ref{cat}). 
        They also add that since images are usually taken in upwards positions, the rotation task is well defined except for round objects which don't have a recognizable orientation. We argue however, that it doesn't only concern round object but any radially symmetric ones (e.g. flowers, some sea creatures), or ones that do not have a consistent shape (e.g. a rock)
        
        Rotation has the same computational cost as supervised learning, making it a good candidate for our project. \cite{rot} train their AlexNet CNN on a single GPU in 2 days.
    
        \subsubsection{Jigsaw}
        
        According to \cite{jigsaw}, one measure of a person's organizational capacity when it comes to visual stimuli, uses Jigsaw puzzles. It thus sounded natural to attempt the same method on CNNs to get them to learn visual representations.
        While the CNN trains on solving a Jigsaw puzzle (guessing the correct locations of the shuffled tiles), it learns parts of the objects along with their spatial positioning at the same time. 
        
        Since similarities between the puzzle tiles don't help in their placement's decision, the focus is put on the differences instead. Additionally, when the location seems to be unclear, having a view on all the tiles helps to determine the appropriate location.
        
        To solve the puzzle, the CNN starts by extracting features using solely the pixels inside a tile. It then uses only those features to decide on the positioning of the tiles. Doing that ensures that the features learnt by the network are discriminative enough.
        At the same time, a number of Jigsaw puzzles (69 on average) are fed to the network at the same time to avoid mapping a certain feature to a position. The tiles also go through configurations for optimal shuffling to make sure that a certain feature can be equally assigned to all of the tiles.

        \subsubsection{Colorization}

        The primary object of \cite{color} was to develop a colorization method that is realistic and has vibrant colors. It managed however, to also set itself apart as a pretext task.
        They achieve nearly photo-realistic results and call their method a "cross-channel encoder" since it's similar to an autoencoder but is inputs and outputs are image channels.
        In order to guess the colors correctly, the cross-channel encoder hints such as textures and scene semantics. They also define their objective as successfully fooling a human rather than guessing the exact color, which might not be possible for some objects (e,g. t-shirt). For that reason, using the ground truth images to validate will not reflect the actual performance of the model. They thus use a "colorization Turing test" where individuals try to spot the colorized image, which they failed at at a 32\% rate. 
    
        Another issue relating to the wide plausible colors that an object might take is that the traditional colorizing technique is by using regression. As a result, those objects end up with grey-like hues since all the different colors that they could take get averaged. What \cite{color} do different is they use colorization as a classification task and each pixel gets a distribution of potential colors.

        \subsubsection{SimCLR}
        
        SimCLR \cite{simclr} is a self-supervised learning framework relying on contrastive loss. The idea behind this loss is to maximize the "agreement" between matching image-label pairs (called positive) and minimize it with all others (negative pairs). 
        
        \cite{simclr} experimented with different image transformations and found the most optimal ones to be random cropping and resizing (necessary resizing since the model is trained on ImageNet\cite{simclr} images which have different sizes) and color distortion, offering a combination of spatial transformation (crop) and appearance transformation (color).
        The color distortion turned out to be crucial since the CNN learns to use color distribution from the pixel intensities histograms to recognize positive pairs, which result in overfitting.
        The transformations are applied in an asymmetric way (applied to only one branch of the network) and although that complicates the predication task, a significant improvement is noticed in the representations quality.

        SimCLR managed to achieve and accuracy that is comparable to a pre-trained Resnet-50. It benefits from large batch sizes and longer training periods since both contribute in it seeing more negative pairs and thus making convergence easier.
        It uses Resnet \cite{simclr} as its base encoder for feature extraction and uses heavy computational resources: 100 epochs on cloud TPUs, with 32-128 cores and huge batch sizes (4000+).


Although SimCLR is the most performant, it requires computational resources that we do not have access to. We're therefore left with the 3 other and rotnet is the one who's most performant (under XY conditions). We have therefore opted for using the Rotation pretext task, which comes second to SimCLR among the presented tasks and relies on hardware similar to the one on our disposition.


We however have decided to use models pre-trained on each one of these pretext tasks (from \url{https://github.com/facebookresearch/vissl/blob/master/MODEL_ZOO.md}) and fine-tune them on our image captioning downstream task. Doing this allows us to assess our model to make sure whether getting a certain result is due to using self supervised learning in image captioning or if it's due to a potential error in our implementation.

\subsection{Architecture}
        \begin{figure}
            \centering
            \includegraphics[width=\textwidth]{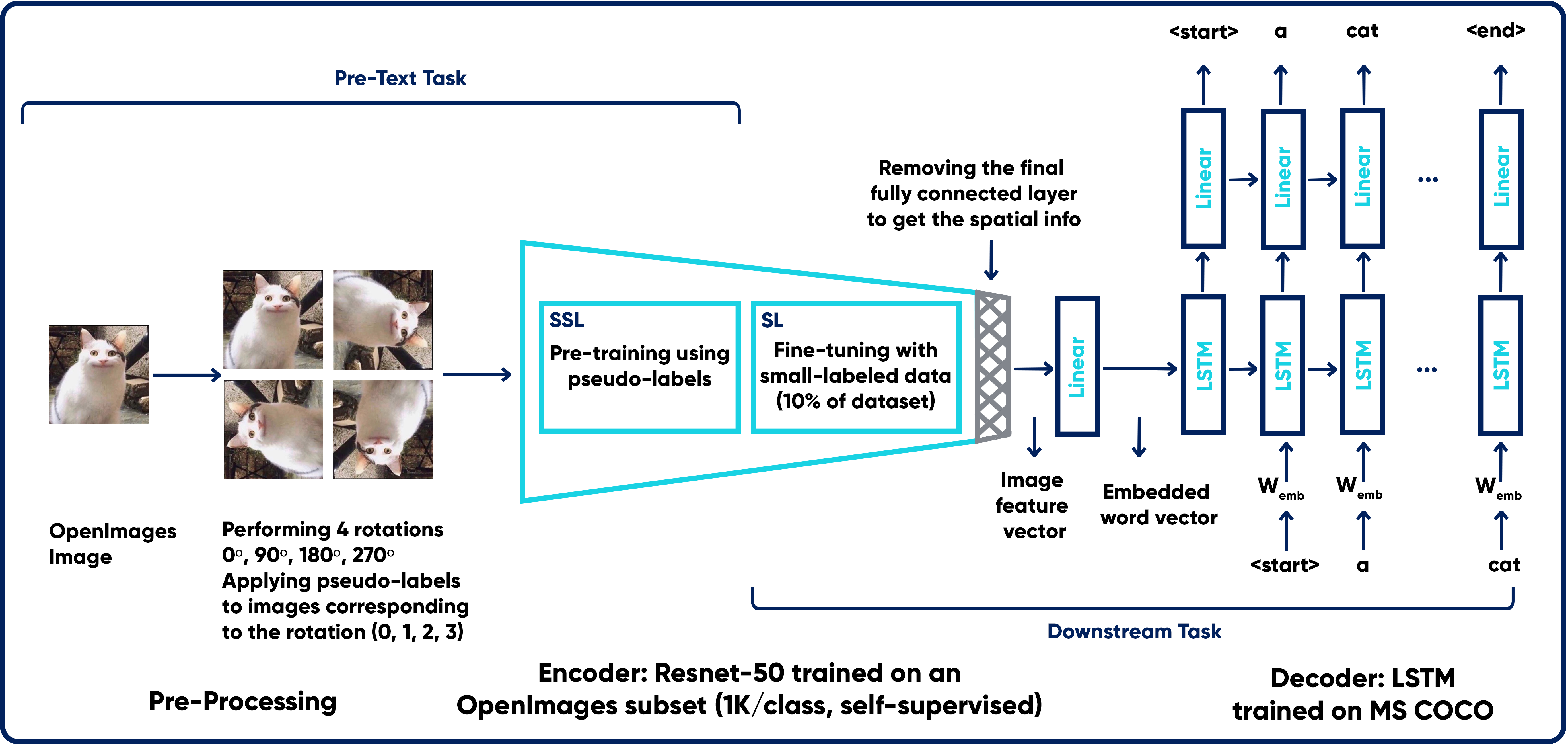}
            \caption{Architecture of our image captioning model}
            \label{arch}
        \end{figure}
        \subsubsection{Pretext Task}
        In the pretext task, we start by performing 4 rotations on the images: 0º, 90º, 180º and 270º. Each of these rotations is given a "pseudo-label" depending on their angle (label 0 to 0º, label 1 to 90º and so on).
        
        
        We choose to use these 4 rotations because they're the only ones that don't lose data when rotated. Rotating an image to 45º for instance, will result in some of its edges being cut off (\ref{cat}).
        
        
            
            Generally, Convolutional Neural Network (CNN) is a type of deep neural network that is most powerful in image processing tasks, such as sorting images into groups. CNN's consist of layers that process visual information. A CNN first takes in an input image and then passes it through these layers. There are a few different types of layers: convolutional, pooling, and fully-connected layers. A classification CNN takes in an input image and outputs a distribution of class scores, from which we can find the most likely class for a given image. 
            
            The encoder we provide extracts features from a batch of pre-processed images using the pre-trained ResNet-50 architecture, without the final fully-connected layer. The output is then transformed to a vector and passed through a Linear layer to make the feature vector of the word embedding's size. The output of a given convolutional layer is a set of feature maps (also called activation maps), which are filtered versions of an original input image.
            
            Convolutional Layer is made of a set of convolutional filters. Each filter extracts a specific kind of feature, ex. a high-pass filter is often used to detect the edge of an object. Pooling Layer takes in an image (usually a filtered image) and output a reduced version of that image. Maxpooling layers look at areas in an input image (like the 4x4 pixel area pictured below) and choose to keep the maximum pixel value in that area, in a new, reduced-size area. Maxpooling is the most common type of pooling layer in CNN's, but there are also other types such as average pooling. A fully-connected layer's job is to connect the input it sees to a desired form of output. Typically, this means converting a matrix of image features into a feature vector whose dimensions are 1xC, where C is the number of classes.

\begin{figure}[h!]
                \centering
                \label{resnet} \includegraphics[width=\textwidth]{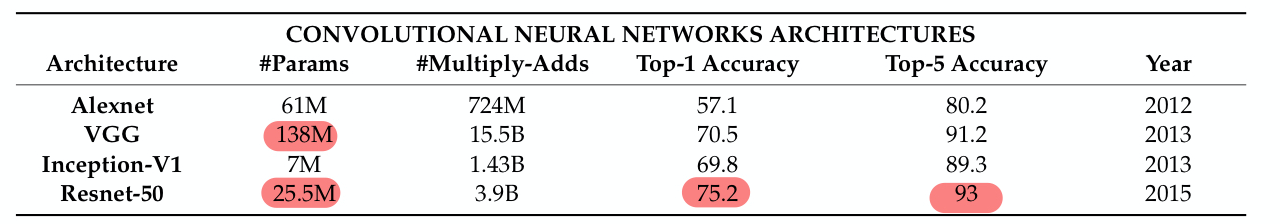}
                \caption{Table of comparison of CNN architectures from \cite{systematic}}
            \end{figure}

            
            It is known in the field of deep learning that better performance can be achieved with deeper networks but that entails facing gradient vanishing and gradient explosion \cite{SSL}.\\ ResNet was suggested to provide a remedy for this problem through using a “skip connection” in convolution blocks, that sends the previous feature map to the following blocks. Since ResNet is smaller and performs better, it’s very common for it to be used as the base network in other image-related tasks \cite{SSL}. ResNet was used in 33 of 78 reviewed articles in \cite{systematic} for its high efficiency in terms of computation compared to the other convolutional networks. It has also shown the best performance in comparison to AlexNet, VGG and Inception-V1 in both Top-1 and Top-5 accuracy according to \cite{systematic} and has less parameters than VGG, saving in computational resources (Fig. \ref{resnet}). 

            Using self-supervised learning, our ResNet-50 encoder learnt the image feature representation by making the model predict image rotations. In the quest of predicting the rotation, the model is able to learn good representation of features from the image. In order to predict how much degree the image is rotated, the model somehow has to look into the image and understand (learns good features to be able to detect their rotation), and from there it detects the whole image rotation. From the proxy task training, the pre-trained model can be transferred (transfer learning) and used on the actual downstream task. The pre-trained model at this stage is with better initialization, after being trained on the proxy task. In our project, we trained on unlabeled subset of OpenImages dataset of 1K images/class. We trained it for 100 epochs with a batch size of 32. We fine-tuned the model afterwards using 10\% of the used subset of OpenImages dataset. For evaluating the pretext task by itself, logistic regression was used. So whatever representations have been learned on the pretext task, we pass that a logistic regression and we perform the classification task to see whether it is doing good or not. Hence, if the representations are good enough, the classifier will be performing good. 

            
            
        
            
            
            
        
            
            
            
            
            
            We decided to use a subset of OpenImages by only using a maximum of 1000 images per class and removing any class that has less than 5 images, resulting in a total of 563 classes. There are, however, only 135 classes that have 1000 or more images. In retrospect, we believe that as a future work it will be better to balance out the dataset more by applying some data augmentation methods or removing the classes below a higher threshold of number of images.
            
            
    
    
        
        
        
        
        

    \subsection{Downstream Task}
        
        In the downstream task, we fine-tune the encoder using supervised learning with small labeled data (only 10\% of the labels are included). We then remove the the final fully connected layer of the CNN that normally gives classification scores in order to get spatial information represented as image feature vectors. These vectors are then passed to the decoder. Our decoder is a special type of recurrent neural networks (RNNs), which is the long short-term memory (LSTM).
        
        
            The recurrent layer of all RNNs has feedback loops. This allows them to keep details in 'memory' for a long time. However, training regular RNNs to solve problems that involve understanding long-term temporal dependencies can be challenging. This is due to the fact that the loss function's gradient decays exponentially over time (called the vanishing gradient problem). \cite{RNN_LSTM} In addition to regular units, LSTM networks use special units. A 'memory cell' in an LSTM unit can store data for a long time. Series of gates are used to control how the information enters and exists the memory and when it is forgotten. They make it learn long-term dependencies using this architecture. 
            
            All inputs are passed as a sequence to an LSTM. A sequence looks like this: first a feature vector that is extracted from an input image, then a start word, then the next word, the next word, and so on. The LSTM is defined such that, as it sequentially looks at inputs, it expects that each individual input in a sequence is of a consistent size and so the feature vector is embedded and each word so that they are of a defined embed size. Linear layer is used at the end to map the hidden state's output dimension. The LSTM cell allows a recurrent system to learn over many time steps without the fear of losing information due to the vanishing gradient problem. It is fully differentiable, therefore gives us the option of easily using backpropagation when updating the weights. So, an LSTM looks at inputs sequentially. In PyTorch, there are two ways to do this. The first is intuitive: for all the inputs in a sequence, which in this case would be a feature from an image, a start word, the next word, the next word, and so on (until the end of a sequence/batch), each input is looped through. The second is when the LSTM is fed with the entire sequence having it producing a set of outputs and the last hidden state. According to the paper reference \cite{show_tell_lessons}, we used the second approach, as it was proven to excel over the first one.

            
            
            
            The Microsoft Common Objects in COntext (MS COCO) dataset is a large-scale dataset for scene understanding. The dataset is commonly used to train and benchmark object detection, segmentation, and captioning algorithms. We used the image-caption dataset to train a CNN-LSTM model to generate automatically captions from images. The project is implemented using PyTorch framework with other python  modules including NLTK, Numpy, Matplotlib. The COCO API for MS COCO dataset was used to retrieve the data. The API is first initialized, then we load the dataset in batches using a data loader. Doing this, we can access the dataset using the dot operator of the data loader variable "data\_loader.dataset", where "dataset" is an instance of the "CoCoDataset" class. 

            Using the data loader, we firstly specify some parameters for the tmages pre-processing: using "transform" argument, we converted the COCO training images into PyTorch tensors before being fed to the CNN encoder. Then we resized the images, cropped them to 224x224 from a random location, randomly flipped the images horizontally, and finally normalized them after converting them to tensors. For the CoCoDataset Class, the \_\_getitem\_\_ method is used to determine how an image-caption pair is pre-processed before joining a batch. The image is pre-processed using the same transform that was provided with when the data loader was instantiated, after it has been loaded into the training folder with the name path. In addition, the captions are pre-processed for training. The model predicts the next token of a sentence from previous tokens, so the caption is converted associated with any image into a list of tokenized words before casting it to a PyTorch tensor that we can use to train the network. The \_\_init\_\_ method is where we intialized an instance of the Vocabulary class with defining different parameters: the vocab threshold, the vocab file path, the start word , the end word, the unknown word and the annotations file paths. Using NLTK Library, tokenization is where all string-valued captions are converted into integers list, and this is before casting these tokenized lists to PyTorch tensors. The tokenization process starts by initializing an empty list and appending an integer as a caption start mark "<start>" (its value is integer 0), and the same happens at the end of a caption with a caption end mark "<end>" (its value is integer 1). After the caption's start mark, we append to the list the integers corresponding to the tokens till the end mark . Finally, the integers list is converted to a PyTorch tensor.
            
            Example:
            
            - Sentence (List of Tokens): 
            [<start>, 'a', 'man', 'riding', 'a', 'bike', 'while', 'listening', 'to', 'music', '.', <end>]
            
            - Corresponding list of integer values:
            [0, 2, 74, 12, 2, 18, 56, 865, 7, 66, 369, 1]

            The length of the captions in the MS COCO dataset varies. To produce batches of training data, we started by sampling a caption length, having the probability that any length is drawn is proportional to the number of captions with that length in the dataset. Then we got a batch of image-caption pairs with a given batch size, where all captions are the same length as the sampled captions. The get\_train\_indices method samples a caption length, and then samples the indices of the defined bacth size corresponding to training data points with captions of that length. The data loader receives these indices and uses them to retrieve the corresponding data points. The batch's pre-processed images and captions are saved to images and captions.

            When we deal with the image captioning task, especially the RNN component, we define some essential hyperparameters. For instance, word embedding size specifies the dimentiontioality of the image and word embedding. Besides, hidden size states the number of features in the hidden state of the RNN decoder. In addition, the vocab threshold defines the minimum word count threshold that each caption should have.

\section{Experiments \& Results}
 
            The training of the CNN-LSTM model is by specifying hyperparameters and setting other options that are important to the training procedure. For instance, we tuned the batch size controlling the image-caption pairs used to update the weights in each step. Besides, we tuned the vocab threshold accepting only words to the vocabulary that exceeds it. In addition, we tuned the hidden size and embed size defining the features number in the LSTM hidden states and dimensionality of the word embeddings and image respectively. 
            
            The first part of the architecture is the pre-trained ResNet CNN. As in the papers referenced \cite{show_tell} \cite{show_attend_tell}, the LSTM decoder follows the described architecture. The paper specifies values for a variety of hyperparameters: a vocab threshold of 4, a embed size of 512 and a hidden size of 512. These hyperparameters have been used here. The paper also mentioned the dropout usage, although the value was not specified. However, in PyTorch this requires at least two hidden layers. With a single layer and no dropout, reasonable results have been got here. The paper didn't specify a batch size, so different batch values have been tried. Moreover, the number of epochs was found to be suitable to be 10 concerning the dataset size and the batches taken out of it for every training epoch. For selecting the optimizer, firstly, SGD is started with as specified on the paper. In fact, it appeared to be slow even with 0.1 learning rate. Afterward, Adam was used and it performed better and faster with learning rate of 0.001 and 0.9 momentum.
            
            Using the same training hyperparameters, we introduced self-supervised learning (SSL) to image captioning using rotation prediction for learning image feature representation. The rotation is used as a pretext task to train the encoder for the image captioning task. It was chosen since it’s initially trained in the original paper on similar conditions (single GPU, 100 epochs, 2 days). Moreover, 4 rotations were performed (0º, 90º, 180º, 270º), which yielded the best results in the original paper. 
            
            In addition, the following pre-trained VISSL models are used to assess our work: Rotnet, SimCLR, Jigsaw and Colorization. They were selected as they were trained by Facebook research on similar number of epochs and hyperparameters\cite{VISSL}.

            PyTorch is definitely a newer framework, but it's fast and intuitive when compared to Tensorflow variables and sessions. PyTorch is designed to look and act like normal Python code: PyTorch neural nets have their layers and feedforward behavior defined in a class. Defining a network in a class means that it is possible to instantiate multiple networks, dynamically change the structure of a model, and these class functions are called during training and testing. PyTorch is also great for testing different model architectures, and the networks are modular, which makes it easy to change a single layer in a network or modify the loss function and see the effect on training \cite{pytorch}.
            
\begin{figure}
    \centering
    \includegraphics[width=\textwidth]{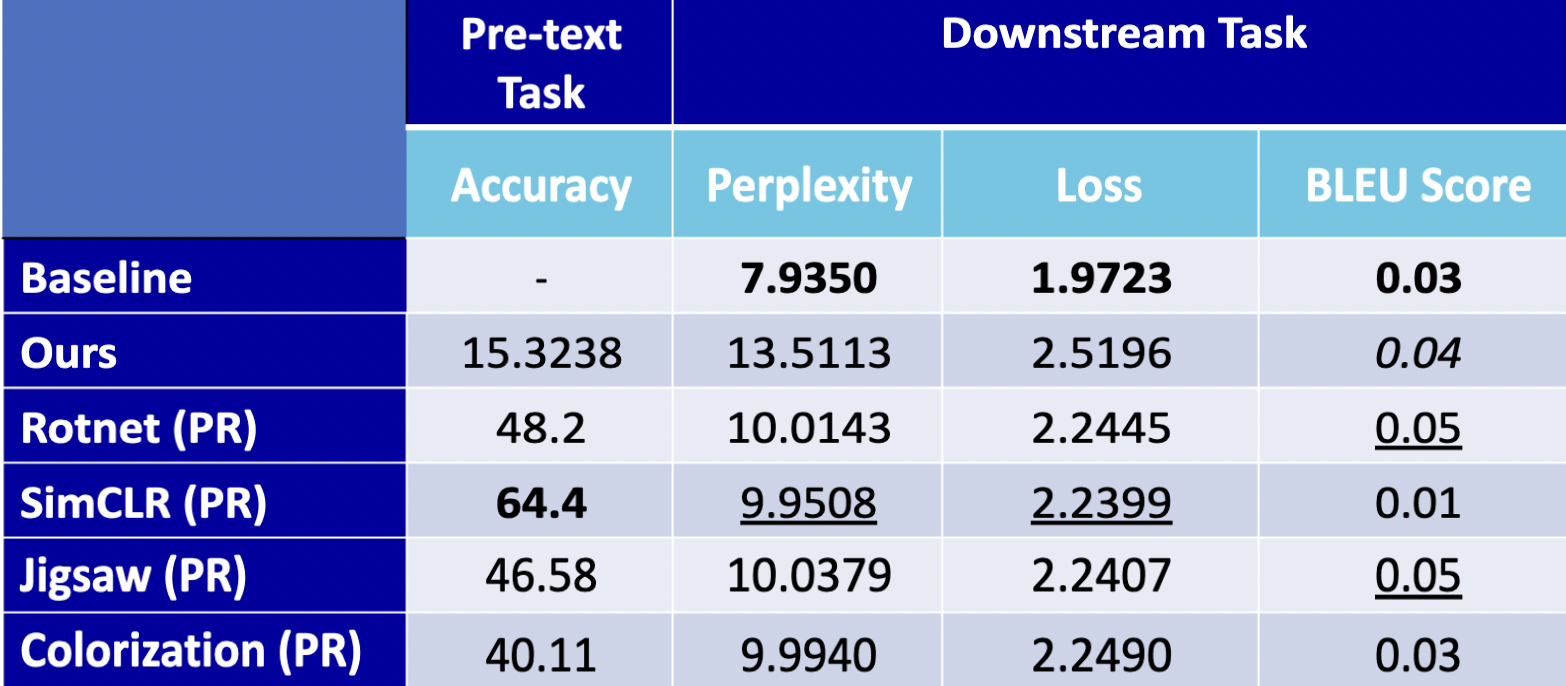}
    \caption{Comparison of accuracy, perplexity and loss between the different models. PR: Pre-trained}
    \label{results}
\end{figure}

        As our optimizer, we use Adam, \cite{adam} since it has the best validation accuracy overtime, with a momentum of 0.9 and a learning rate of 0.001. We chose those values as their combination with the optimizer minimizes the training time. We train our encoder and decoder for 100 and 10 epochs and with a batch size of 32 and 64 respectively.
        
        The results were surprising to us as we expected SimCLR to be the most performant model after the baseline on the downstream task. (Fig. \ref{results}) shows, however, that although SimCLR has a considerably higher accuracy when it comes to the pretext task, it ends up being comparable to the other pre-trained models in both perplexity and loss.
        
\begin{figure}
    \centering
    \includegraphics[width=0.7\textwidth]{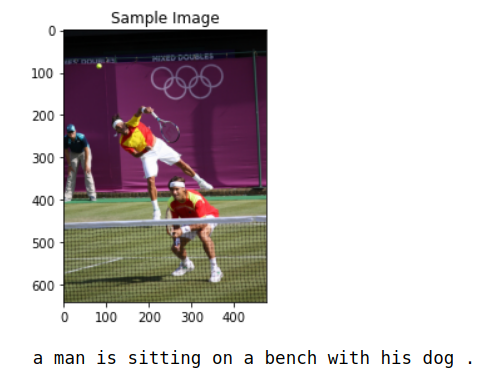}
    \caption{Caption from our SSL model}
    \label{caption_good}
\end{figure}

\begin{figure}
    \centering
    \includegraphics[width=0.7\textwidth]{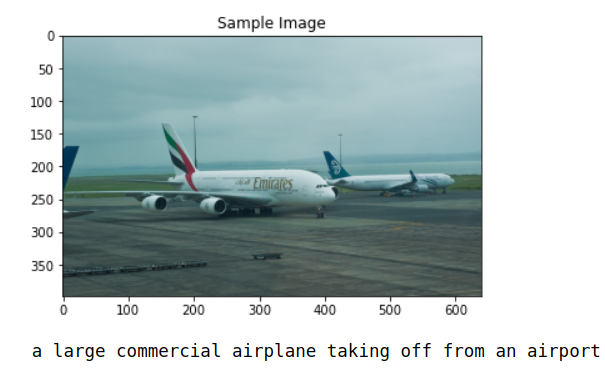}
    \caption{Caption from the baseline}
    \label{caption_bad}
\end{figure}

        The BLEU score however, is interesting. All of the models score close to each other and around the baseline. We explain a high loss and high perplexity (negative) and a low BLEU score (positive) by looking at the captions in (Fig. \ref{caption_good}) and (Fig. \ref{caption_bad}). One of them describes the image correctly, resulting in a low loss, perplexity and BLEU score. (Fig. \ref{caption_bad}) however, has a caption that doesn't match the image, making its loss and perplexity high but the BLEU score being low has to do with it being a correct sentence that could have been generated by a human. 
        If we were to compare the captions without their paired images, we wouldn't see a difference between their quality. 
        
    We therefore conclude that having high classification accuracy doesn't always guarantee accurate captions, as demonstrated by the SimCLR performance on the downstream task. We think that SimCLR could potentially perform better if we provide it with the hardware that it needs.
   
    We can also see how the BLEU score is only assessing 1 aspect of the captions and that having multiple metrics to evaluate your model give you a fuller picture of its strengths and limitations.
    
    Our next work is to explore other approaches with image captioning, as the rotation prediction method did not yield satisfactory results, even though it sounded like it could potentially work in theory.
\bibliographystyle{plain}
\bibliography{references}
\end{document}